\documentclass{article}

\usepackage{graphicx}
\usepackage{amsthm,amsmath,amsfonts}
\usepackage[colorlinks,linkcolor=red,anchorcolor=blue,citecolor=blue]{hyperref}
\usepackage{comment}
\usepackage{stmaryrd}
\usepackage[dvipsnames]{xcolor}
\usepackage{algorithm}
\usepackage{algorithmic}
\usepackage{smile}
\newcommand{\R}{\mathbb{R}}
\newcommand{\E}{\mathbb{E}}

\newcommand{\KL}{\mathrm{KL}}

\newcommand{\politex}{\textsc{Politex}\;}

\def\botao{\color{purple}}
\def\KL{\text{KL}}

\usepackage{hyperref}

\usepackage[accepted]{icml2021}

\icmltitlerunning{Optimization Issues in KL-Constrained Approximate Policy Iteration}

\begin{document}

\twocolumn[
\icmltitle{Optimization Issues in KL-Constrained Approximate Policy Iteration}



\icmlsetsymbol{equal}{*}

\begin{icmlauthorlist}
\icmlauthor{Nevena Lazi\'c}{dm}
\icmlauthor{Botao Hao}{dm}
\icmlauthor{Yasin Abbasi-Yadkori}{dm}
\icmlauthor{Dale Schuurmans}{goo,ua}
\icmlauthor{Csaba Szepesv\'ari}{dm,ua}
\end{icmlauthorlist}

\icmlaffiliation{dm}{DeepMind}
\icmlaffiliation{goo}{Google}
\icmlaffiliation{ua}{University of Alberta}

\icmlcorrespondingauthor{Nevena Lazic}{nevena@google.com}


\vskip 0.3in
]




\begin{abstract}
Many reinforcement learning algorithms can be seen as versions of approximate policy iteration (API). While standard API often performs poorly, it has been shown that learning can be stabilized by regularizing each policy update by the KL-divergence to the previous policy. 
Popular practical algorithms such as TRPO, MPO, and VMPO replace regularization by a constraint on KL-divergence of consecutive policies, arguing that this is easier to implement and tune. 
In this work, we study this implementation choice in more detail. We compare the use of KL divergence as a constraint vs. as a regularizer, and point out several optimization issues with the widely-used constrained approach. We show that the constrained algorithm is not guaranteed to converge even on simple problem instances where the constrained problem can be solved exactly, and in fact incurs linear expected regret. With approximate implementation using softmax policies, we show that regularization can improve the optimization landscape of the original objective.  We demonstrate these issues empirically on several bandit and RL environments. 
\end{abstract}

\section{Introduction}

Model-free reinforcement learning (RL) algorithms combined with value function approximation have recently achieved impressive performance in a variety of application domains. 
Many practical algorithms can be viewed as variants of approximate policy iteration (API). API alternates between a \emph{policy evaluation} (PE) step, where one estimates the advantage function of the current policy, and a \emph{policy improvement} (PI) step, where the next policy is obtained by maximizing the expected advantage. 

While standard API often performs poorly (it is not guaranteed to converge and often oscillates in practice, see \citet{bertsekas2011approximate}), several works have shown that learning can be stabilized by regularizing each policy update by the KL divergence to the previous policy. This regularization choice has been theoretically justified 
from the perspective of regret analysis \citep{politex, hao2020provably,shani2020optimistic, cai2020provably}, convergence \citep{shani2020adaptive}, as well as an analysis of error propagation in approximate dynamic programming \citep{vieillard2020momentum, vieillard2020leverage}. 
The minimizer of the KL-regularized objective has an analytic form. However performing the exact analytic policy update is typically inefficient in terms of memory and computation when using neural network function approximation, as it requires storing the advantage functions of all past policies and evaluating them at each step. Practical (and possibly inexact) implementations can be obtained simply by adding a regularization term to the policy optimization objective. 

Rather than using KL regularization, many popular practical versions of API \emph{constrain} consecutive policies to be close in terms of KL divergence.
Examples of such algorithms include TRPO \citep{schulman2015trust}, PPO \citep{schulman2017proximal}, MPO \citep{abdolmaleki2018maximum}, VMPO \citep{song2019v}, and CPO \cite{achiam2017constrained}.
Such constrained policy updates are often described as practical implementations of theoretically justified algorithms that are less conservative \cite{schulman2015trust} or easier to tune \cite{abdolmaleki2018maximum}. Conversely, the empirical success of KL-constrained algorithms is sometimes attributed to similarities to mirror descent \cite{shani2020adaptive}.

In this work, we take a deeper look at the discrepancy between popular practical implementations and their theoretical counterparts. We focus on the following simple question:
\begin{center}
    \emph{Should we use KL divergence as a regularizer or as a constraint in approximate policy iteration?}
\end{center}
We show that from an optimization perspective, the difference between these two implementation choices is highly non-trivial. In particular, we show analytically that the constrained version is not guaranteed to converge even on very simple (bandit) problem instances where all algorithms can be implemented exactly. 
Intuitively, whenever the advantage estimate is noisy such that the estimated best action is different from the true best action, TRPO will move in the wrong direction until it hits the constraint, and produce a suboptimal policy. Since this can happen with positive probability even in a simple multi-armed bandit problem, TRPO will also have linear expected regret. On the other hand, the regularized update results in the noise being averaged out over iterations and obtains sublinear regret on this problem.


In the case where exact optimization is not possible, we describe several efficient implementations of the constrained and regularized problems, and show that regularization can improve the optimization landscape for softmax-parameterized policies. 
We demonstrate these optimization issues empirically on several environments. 



\section{Preliminaries}


 We consider an infinite horizon, undiscounted MDP characterized by $(\cX, \cA, r, P)$, where $\cX$ is the finite state space,  $\cA$ is a finite action space, $r:\cX\times \cA\to[0, 1]$ is the unknown reward function, and $P:\cX\times \cA \to \Delta_{\cX}$ is the unknown transition probability function. 
 We assume that the MDP is weakly communicating, which is a necessary assumption for learning with low regret in this setting \cite{bartlett2012regal}. 
 We define a (stationary) policy as a function $\pi:\cX\to \Delta_{\cA}$ that maps states to distributions over actions. A nonstationary policy is a sequence of maps from histories to probability distributions over actions. Under the weakly-communicating assumption, the expected average reward of policy $\pi$ is not a function of the initial state, and  defined as 
\begin{equation*}
   J_{\pi}:=\lim_{T\to\infty}\mathbb E^\pi \left[\frac{1}{T}\sum_{t=1}^Tr(x_t, a_t)\right],
\end{equation*}
 where $a_t\sim \pi(\cdot|x_t)$ and $x_{t+1}\sim P(\cdot|x_t, a_t)$. 
Let $J_* = \max_{\pi}J_{\pi}$. A policy $\pi$ is said to be optimal if $J_{\pi} = J^*$.

The value function of a policy $\pi$ is defined as:
$$
V_{\pi}(x) = \mathbb E^{\pi}\Big[\sum_{t=1}^{\infty}r(x_t, a_t) - J_{\pi}|x_1 = x\Big].
$$
The state-action value function $Q_{\pi}(x,a)$ is the (unique up to a constant) solution to the following Bellman equation:
\begin{equation}\label{eqn:Bellman_eqn}
     Q_{\pi}(x,a) = r(x,a) -J_{\pi} + \sum_{x'}P(x'|x, a)V_{\pi}(x').
\end{equation}
The advantage function of a policy $\pi$ is defined as $A_\pi(x, a) = Q_\pi(x, a) - V_\pi(x)$. We will sometimes use the notation $A_{\pi}(x, \pi') =\E_{a \sim \pi'(\cdot|x)}[A_\pi(x, a)]$ for both advantage functions and action-value functions. Note that $Q_\pi(x, \pi) = V_\pi(x)$.

Let $\mu_\pi$ denote the stationary state distribution of a policy $\pi$, satisfying $\mu_\pi(x') = \E_{x\sim \mu, a \sim \pi}[P(\cdot|x, a)]$. In weakly-communicating MDPs, $\mu_\pi$ is well-defined and independent of the initial state. We will sometimes write $\mu_\pi$ as a vector, and use $\nu_\pi = \mu_\pi \otimes \pi$ to denote the stationary state-action distribution. 

The regret of an algorithm is
defined as
\begin{equation}\label{def:regret}
    R_T= \sum_{t=1}^T \Big(J_{*} - r(x_t, a_t)\Big).
\end{equation}
In the online learning setting, the learning goal is to find an algorithm that minimizes the  regret $R_T$. 

\section{Related work}
\label{sec:api}

\begin{algorithm}[t!]
\caption{Approximate policy iteration schema}
\begin{algorithmic}[1]\label{alg:api}
\STATE \textbf{Input:} phase length $\tau$, num. phases $K$,  parameter $\eta$
\STATE \textbf{Initialize:} $\pi_1(a|x) = 1 / |\mathcal{A}|$ $ \forall x, a$;
\FOR{$k=1,\ldots, K$}
\STATE $\cD_k = \text{CollectData}(\pi_k, \tau)$ 
\STATE $\widehat A_{\pi_k} = \text{PolicyEvaluation}(\cD_k)$ 
\vskip -0.2cm
\STATE $\pi_{k+1} = \text{PolicyImprovement}(\cD_k, \widehat A_{\pi_k}, \pi_k, \eta)$
\ENDFOR
\STATE \textbf{Output:} $\pi_{K+1}$
	\end{algorithmic}
\end{algorithm} 

In this section, we survey some of the existing works on regularized approximate policy iteration (API).
API alternates between a \emph{policy evaluation} step, and a \emph{policy improvement} step, as shown in the schema in Algorithm~\ref{alg:api}. 
 During policy evaluation in iteration $k$, the agent executes the current policy $\pi_k$ for $\tau$ steps, and computes an estimate of either the action-value function $Q_{\pi_k}$ or the advantage function $A_{\pi_k}$.  In the policy improvement step, the next policy is set to be greedy w.r.t. the estimate $\widehat A_{\pi_k}$. 
When the action-value functions are exact, policy iteration is guaranteed to converge to the optimal policy. However, with approximation error, policy iteration tends to perform poorly - it is not guaranteed to converge, and often oscillates in practice \cite{bertsekas2011approximate}.

\subsection{Constrained policy updates}

One of the first works to 
improve the monotonicity of API updates
is the 
Conservative Policy Iteration (CPI) algorithm of \citet{kakade2002approximately}. CPI sets each policy to be a mixture $\pi_{k+1} = (1-\alpha)\pi_k + \alpha \pi_k^{gr}$, 
where the $\pi_k^{gr}$ maximizes the approximate expected advantage objective:
\begin{align}
\label{eq:exp_adv}
    L_{\pi_k}(\pi) = \E_{s \sim \mu_{k}, a \sim \pi(\cdot|s)}[\widehat A_{\pi_k}(s, a)] \,.
\end{align}
Here $\mu_k$ is the empirical state distribution obtained by executing the policy $\pi_k$.
For a particular choice of the mixture coefficient $\alpha$, each policy produced by CPI improves performance over the preceding policy with high probability. 
Unfortunately, this algorithm is impractical with neural network function approximation as it requires storing and evaluating a large number of policies. 
 
The Trust Region Policy Optimization (TRPO) algorithm \cite{schulman2015trust} is motivated by CPI, but makes a number of approximations.  The authors show that similar monotonic improvement can be obtained by (greedily) optimizing  \eqref{eq:exp_adv} combined with a KL divergence regularization term.
However, arguing that the constrained updates may be too conservative, they replace the regularized objective with a constrained optimization problem:
\begin{align}
\label{eqn:TRPO}
    \max_\theta \; L_{\pi_k}(\pi_\theta) \;\;\; {\rm s.t.} \; \E_{x \sim \mu_k} \left[ D_{KL}(\pi_k(\cdot|x) \parallel \pi_\theta(\cdot|x) \right] \leq \eta
\end{align}
where $\theta$ parameterizes the next policy $\pi_{k+1}$. 
Here the KL constraint radius $\eta$ is typically a tuned parameter.
There exist different approaches to approximately solving the above optimization problem \cite{schulman2015trust,achiam2017constrained,song2019v}, and the authors propose a conjugate gradient method with a backtracking line search.
 PPO \cite{schulman2017proximal} optimizes a clipped importance-weighted version of the expected-advantage objective, and is motivated as a first-order approximation of CPI and TRPO. 

Another line of KL-constrained policy iteration algorithms includes MPO \cite{abdolmaleki2018maximum} and VMPO \cite{song2019v}. These algorithms motivate their updates by an analogy to the EM-algorithm, where policy improvement corresponds to the M-step. The proposed theory suggests to update policies by optimizing a KL-regularized objective. However, the authors choose to replace the regularization with a KL constraint in their practical implementation, arguing that that the corresponding parameter is easier to tune. 

\subsection{Regularized policy updates}

More recently, several works have proposed regularizing policy updates by KL-divergence to the previous policy 
\citep{politex, hao2020provably, vieillard2020leverage, vieillard2020momentum,tomar2020mirror}. As a concrete instantiation, the \politex algorithm \cite{politex} updates policies as
\begin{align*}
&  \pi_{k+1}(\cdot|x) = \argmax_{\pi \in \Delta_{\cA}} \widehat A_{\pi_k}(x, \pi) - \eta^{-1} D_{\KL}(\pi \parallel \pi_k(\cdot|x)) 
\end{align*}
From a theoretical perspective, this update can be seen as running mirror descent (MD) in each state $x$, with negative entropy regularization and using advantage functions $\widehat A_{\pi_k}(x, \cdot)$ as losses.\footnote{Note that the above policy update remains the same if we replace advantage functions $\widehat A_{\pi_k}$ with action-value functions $\widehat Q_{\pi_k}$.} When the advantage estimation error is sufficiently small and scales as $O(1/\sqrt{\tau})$, this algorithm has a sublinear regret guarantee of $O(T^{3/4})$. The regularized update is also theoretically justified from the perspective of error propagation in approximate dynamic programming; in particular \citet{vieillard2020leverage} show that it results in errors being averaged over iterations. 

The KL-regularized policy improvement step has an analytic solution of the following form:
\begin{align}
& \pi_{k+1}(\cdot|x)  \propto \exp\bigg(\eta \sum_{i=1}^k \widehat A_{\pi_i}(x, \cdot)\bigg) \,.  \label{eq:politex}
\end{align}
Unfortunately, if advantage functions are approximated by neural networks, the above update requires us to store the parameters of all past networks in memory, which is impractical. Possible heuristics for ensuring memory efficiency include subsampling action-value networks \citep{politex} and/or distillation to approximate the sum by a single network \citep{vieillard2020leverage}. 

Another memory-efficient option is to use a parameterized policy $\pi_\theta(\cdot|x)$, and optimize the KL-regularized objective w.r.t. parameters $\theta$ using gradient descent on data from the most recent iteration, as suggested in the Mirror Descent Policy Optimization (MDPO) algorithm of  \citet{tomar2020mirror}:
\begin{align}
  \max_\theta  
   \; L_{\pi_k}(\pi_\theta) - \eta^{-1}\E_{x \sim \mu_k} \left[ D_{KL}(\pi_\theta(\cdot|x) \parallel \pi_k(\cdot|x)) \right] \,. \label{eq:mdpo}
\end{align}
Superficially,  \eqref{eq:mdpo} is quite similar to the TRPO update \eqref{eqn:TRPO}. In fact, the regularized (mirror descent) updates are sometimes referred to ``exact TRPO" \cite{neu2017unified} and used as a theoretical justification of TRPO \cite{neu2017unified,shani2020adaptive}. 
None of the previous works analyze the effects of approximating the regularized objective by a constrained optimization problem. As we will show, the difference between these two updates is highly non-trivial. Most strikingly, in a simple case two-armed bandit environment where both updates can be implemented exactly, the constrained version is not guaranteed to converge when the advantage estimates are noisy. For this setting, we show that the expected regret of the constrained implementation is linear.
\section{Convergence and regret}


In this section, we show that TRPO is not guaranteed to converge even on a simple two-arm stochastic bandit problem (an MDP with a single state), where we can implement the constrained update exactly.  Furthermore, we show that the expected regret of TRPO on this problem has a linear lower bound. On the other hand, the regularized version inherits the theoretical guarantees of mirror descent \citep{shani2020adaptive}. 

Consider a two-armed stochastic bandit problem, where the expected rewards for the two arms are $r(0) =-\Delta / 2$ and $r(1) = \Delta / 2$. At each pull $a_t$, the learner observes $r_t = r(a_t) + z_t$, where $z_t \sim \mathcal{N}(0, \sigma^2)$.
Following the API learning schema in Algorithm \ref{alg:api}, we run current policy for $\tau$ steps within each phase, estimate the mean reward from the collected data using empirical means, and update the policies. 
Let $\widehat \Delta_k = \widehat r(1) - \widehat r(0)$ be the estimate of the reward gap computed using data in phase $k$, where $\widehat r(0), \widehat r(1)$ are empirical reward means (set to 0 if arm $i$ is not pulled). 
Mirror descent updates policies in the simplex as
\begin{equation*}
    \begin{split}
        \pi_{k+1} &=\argmax_{\pi}\Big(\hat{\Delta}_k\pi(1) -\eta^{-1} D_{\text{KL}}(\pi ||\pi_k)\Big)\\
        &= \frac{\exp(\eta \sum_{i=1}^k \widehat \Delta_i)}{1 + \exp(\eta \sum_{i=1}^k \hat{\Delta}_i)} \,.
    \end{split}
\end{equation*}
TRPO updates policies by solving the following scalar optimization problem at each iteration:
\begin{equation}\label{eq:trpo2}
    \begin{split}
         \pi_{k+1} & = \; \argmax_{\pi}  \; \widehat \Delta_k \pi(1) \\
    & \text{s.t.} \; D_{\text{KL}}(\pi_k||\pi) \leq \eta \,. 
    \end{split}
\end{equation}

In this special case, the TRPO update can be computed exactly as the above problem is convex in $\theta$.

We note that even on this simple example, TRPO will not converge to the optimal solution $\pi(1) = 1$ for a fixed $\eta$. Informally, the reason for this is that whenever the empirical reward estimate $\widehat \Delta_k < 0$ (which can happen with positive probability due to noise), the policy parameter $p$ will move in the wrong direction until it hits the constraint, and we will have $\pi_{k+1}(1) \leq \pi_k(1)$. 
We state this more formally in the following lemma.


\begin{lemma}
\label{lemma:convergence}
We define a class of $\delta$-nearly optimal policy $\Pi(\delta)$ such that for any $\pi\in\Pi(\delta)$, we have $1-\delta<\pi(1)\leq 1$. For any fixed radius of KL-divergence $\eta$ and any $\pi_k\in\Pi(\sqrt{\eta/8})$, there always exists a problem instance $\cG = (\Delta, \sigma)$ such that the policy does not improve: 
$$
\mathbb E[\pi_{k+1}(1)|\pi_k(1)=\theta_k]\leq \theta_k.
$$
\end{lemma}

The proof is given in the supplementary material.

Define the expected regret for the described two-armed bandit problem as:
\[\E[R_T] = \E \left[ \sum_{t=1}^T 
\frac{\Delta}{2} - r_t \right] ,
\]
We now show that the expected regret of TRPO on this problem has a linear lower bound. 

\begin{lemma}[Linear regret]
\label{lemma:trpo_regret}
Let $\Phi(\cdot)$ be the standard Gaussian cdf. The expected regret of TRPO with constraint parameter $\eta$ on any problem instance $\cG = (\Delta, \sigma)$ is lower-bounded as 
\[\E[R_T] \geq  \frac{ \Delta}{2} \Phi \left(-\frac{\sqrt{\tau}\Delta}{2 \sigma} \right)(1 - \exp(-\eta)) (T-\tau).\]
\end{lemma}
The proof is again given in the supplementary material. Intuitively, the probability of the "estimation failure" event that $\Delta_k < 0$ in any phase $k$ is at least $\Phi \big(-\frac{\sqrt{\tau}\Delta}{2 \sigma} \big)$ regardless of the policy or the data collection strategy. Whenever the failure occurs, the probability of the next policy taking the optimal action is upper-bounded by $\pi_{k+1}(1) \leq \exp(-\eta) < 1$, resulting in linear expected regret. On the other hand, the regret of mirror descent will be upper-bounded by $O(T^{3/4})$ following the analysis of  \citet{politex}, and the bound can be improved to $O(\sqrt{T})$ if we use all data (rather than just current-phase data) to estimate $\Delta_k$ in each phase $k$.

\begin{figure}[!t]
\centering
\includegraphics[width=0.9\linewidth, trim=1cm 0.1cm 1cm 0.5cm, clip]{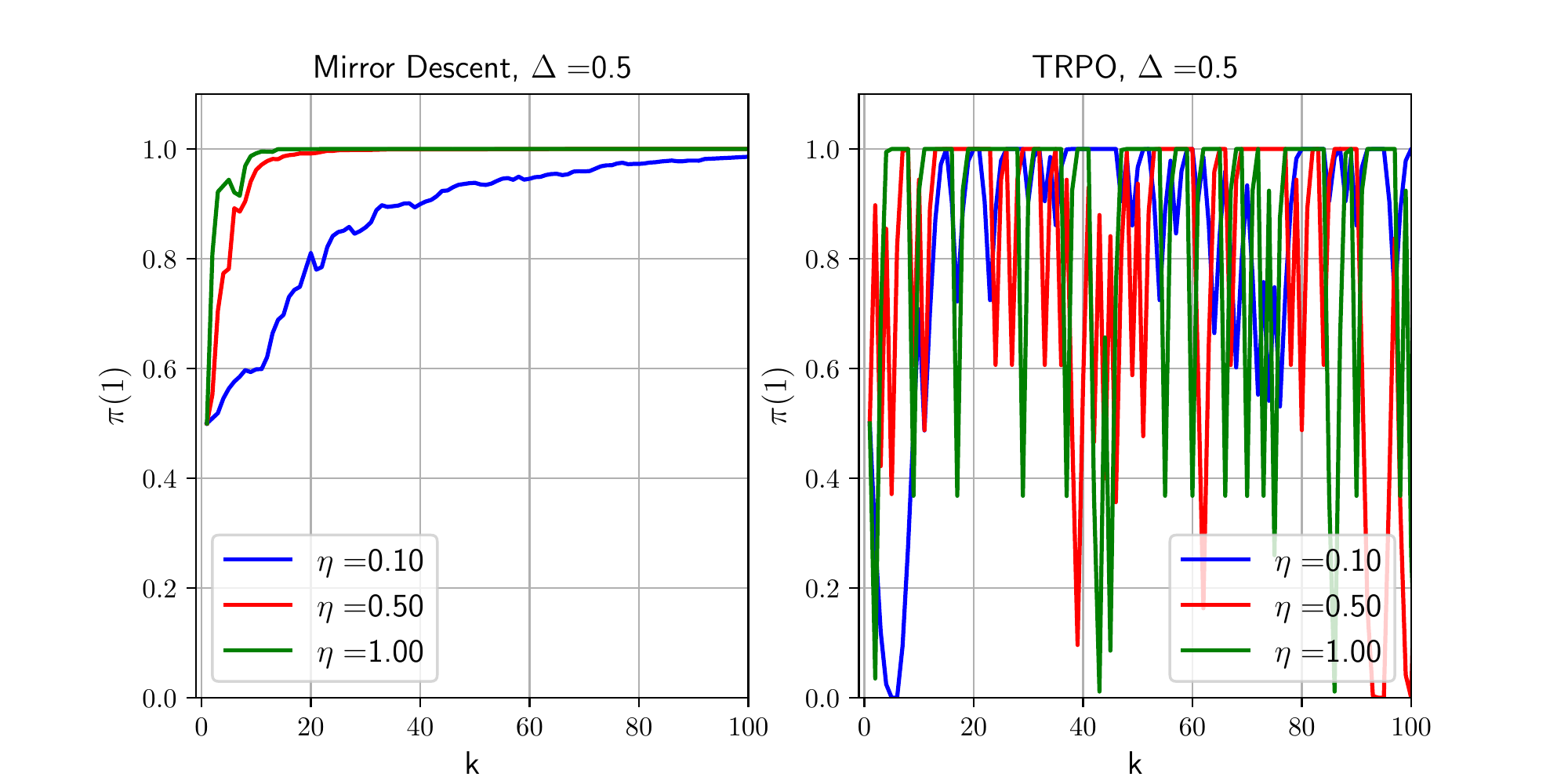}
\includegraphics[width=0.9\linewidth, trim=1cm 0.1cm 1cm 0.5cm, clip]{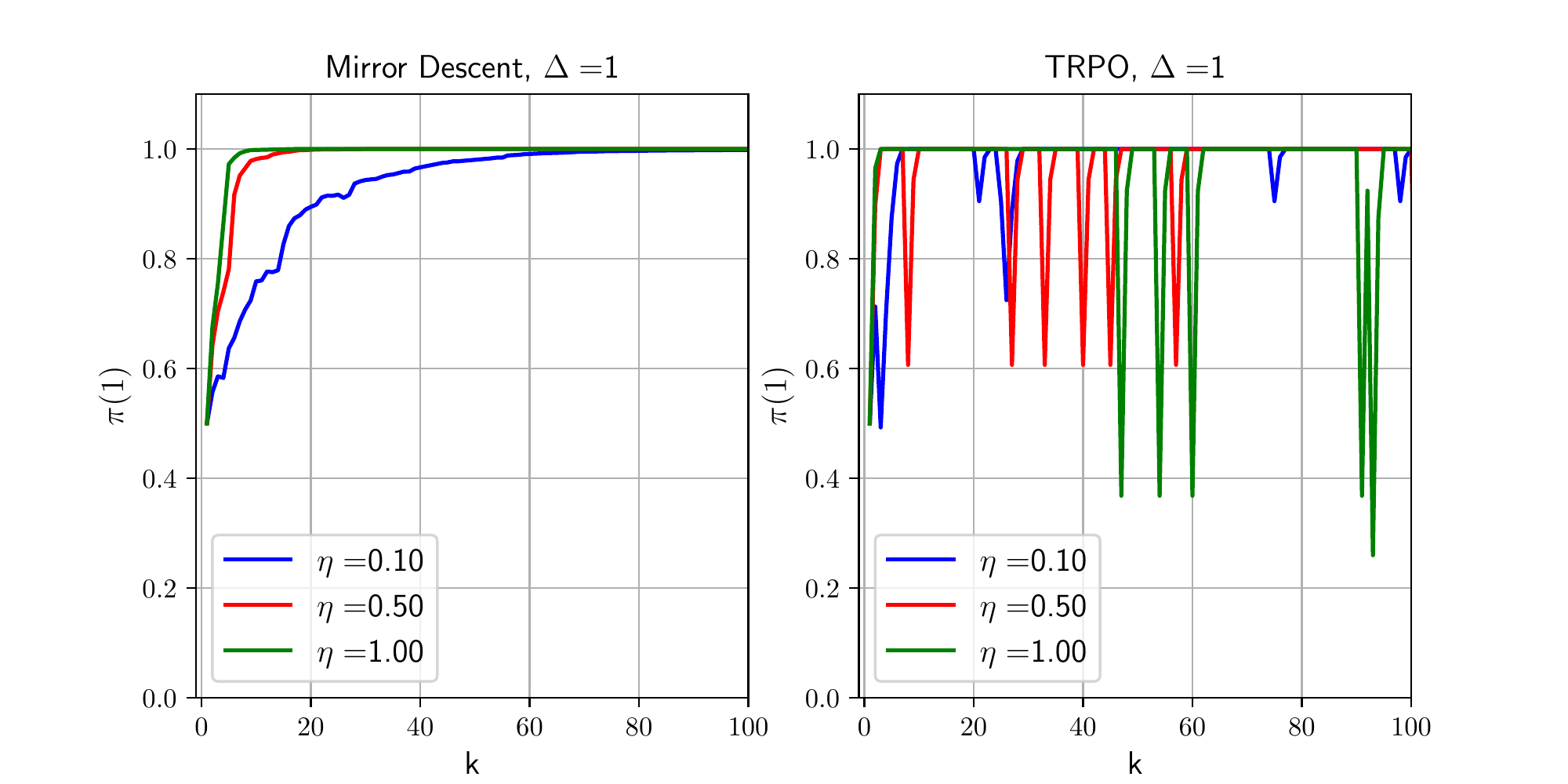}
\caption{Example runs of mirror descent and TRPO on a two-armed stochastic bandit problem with reward gap $\Delta$, demonstrating that TRPO does not converge with noisy rewards.}
\label{fig:trpo}
\end{figure}

We show example runs of the two algorithms on a two-armed bandit problem with $\Delta \in \{0.5, 1\}$, $\sigma^2=1$, $\eta \in \{0.1, 0.5, 1.0\}$, $\tau=20$, and $K=100$ in Figure~\ref{fig:trpo}.  Unsurprisingly, the experiment demonstrates that mirror descent converges to the optimal policy for all problem instances, as the reward noise gets averaged out over iterations. On the other hand, TRPO tends to oscillate even in the "high SNR" case  $\Delta=1$. Although TRPO finds the optimal policy quickly, it switches to a suboptimal policy whenever $\widehat \Delta_k < 0$, which happens with positive probability due to noise. Thus, the empirical success of TRPO can be attributed at least in part to good advantage estimation, and perhaps also to picking the best-performing policy over all iterations.

\section{Optimization landscape}

\begin{figure*}[!ht]
 \centering
\includegraphics[width=0.9\linewidth, trim=0.1cm 0cm 2.2cm 0cm, clip]{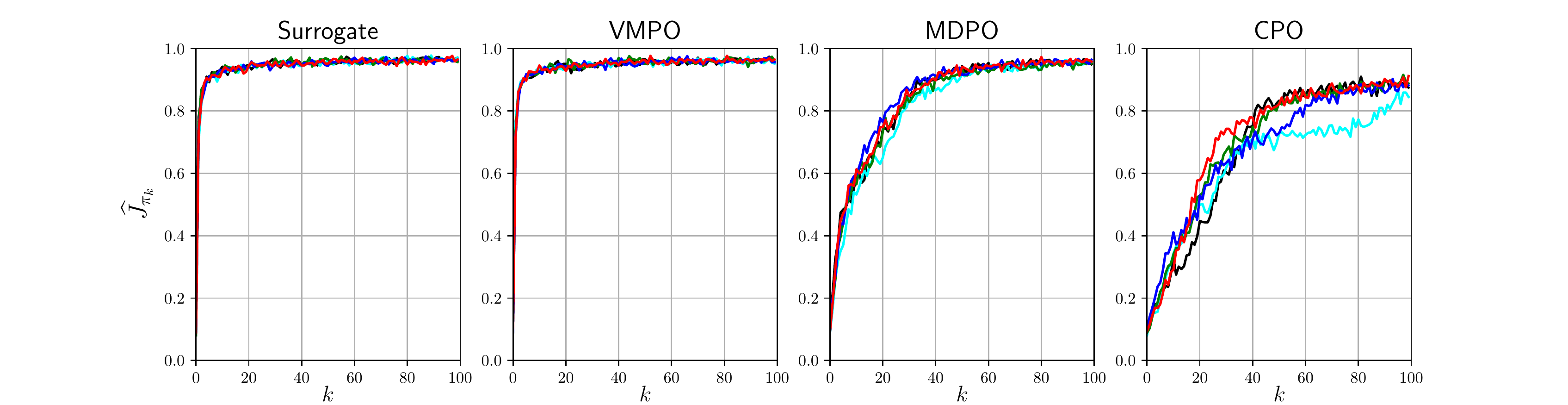}
\caption{Evaluation on the MNIST contextual bandit environment for  $\tau=1000$, showing five random runs for each algorithm.}
\label{fig:mnist}
\end{figure*}




In general, with rich function approximation, the constrained policy improvement step cannot be implemented exactly, while exact implementation of the regularized step as in \ref{eq:politex} is inefficient in terms of memory and computation when using neural networks. In this section, we describe several efficient (but inexact) actor-critic implementations of the two approaches, similar to existing literature. We subsequently survey some of the previous work on the optimization landscape of these problems with softmax-parameterized policies. We observe that the expected-advantage objective optimized in the constrained policy improvement step can exhibit suboptimal plateaus and exponentially many local optima in the worst case even for log-linear policies \cite{surrogate}, and that the landscape can be partially improved by regularization.

\subsection{Algorithms}

\paragraph{Constrained policy optimization (CPO).} At each iteration, we minimize the negative expected advantage:
\begin{align*}
 L_{\pi_k}(\pi_\theta) = -\E_{x \sim \mu_k, a \sim \pi_\theta(\cdot|x)} \left[ 
\widehat A_{\pi_k}(x, a) \right],
\end{align*}
where $\mu_k$ is the empirical distribution corresponding to the $k^{th}$ dataset $\cD_k$. 
We update the objective using batch gradient descent (with gradients computed using the policy gradient trick \cite{sutton2000policy}), check the empirical KL constraint \[\E_{x \sim \mu_k}[ D_{KL}(\pi_k(\cdot|x) \parallel \pi_\theta(\cdot|x)) ]\leq \eta,
\]
after each step, and stop when the constraint is reached.  
This is the simplest possible implementation (though not the most computationally efficient one), and directly comparable to regularized approaches.

\paragraph{MDPO.} 
We minimize $ L_{\pi_k}(\pi_\theta)$ plus a KL regularization term using gradient descent without imposing any constraints, similarly to \citet{tomar2020mirror}:
\begin{align*}
L_{MDPO}(\pi_\theta) :=&  -L_{\pi_k}(\pi_\theta) \\
& + \eta^{-1} \E_{x \sim \mu_k} [ D_{KL}(\pi_\theta(\cdot|x) \parallel \pi_k(\cdot|x)].
\end{align*}

\paragraph{Surrogate loss.}
Let $\psi_k(x, a)$ be a distribution such that  
\[\psi_k(x, a) \propto \pi_k(a|x)\exp\left(\eta \widehat A_{\pi_k}(x, a)\right).
\]
Note that the MDPO objective can be alternatively written as 
$L_{MDPO}(\pi_\theta) = \E_{x \sim \mu_k} [D_{KL}(\pi_\theta(\cdot|x) \parallel \psi_k(x, \cdot))]$ up to a constant.
We can alternatively minimize the following upper bound on $L_{MDPO}$ shown by \citet{norouzi2016reward}, which reverses the KL divergence:
\begin{align*}
   & L_{surr}(\pi_\theta) = E_{x \sim \mu_k} \bigg[ D_{KL}(\psi_k(x, \cdot) \parallel \pi_\theta(\cdot|x)) \\
    & + 0.25 \big\| q_k(x, \cdot) + \eta \widehat A_{\pi_k}(x, \cdot) - q_{\theta}(x, \cdot) - v(x){\bf 1} \big\|_2^2\bigg].
\end{align*}
Here $\pi_\theta \propto \exp(q_\theta)$, $\pi_k \propto \exp(q_k)$, and $v(x)$ is an arbitrary scalar baseline that can be optimized for each example. 
As shown in \citet{surrogate}, $L_{surr}$ is an upper bound on $L_{MDPO}$ that is strongly convex in $q_\theta$. 

\paragraph{VMPO.} 
The VMPO algorithm \cite{song2019v} performs policy improvement by optimizing only the KL component of $L_{surr}$, subject to yet another KL constraint. We will refer by VMPO to optimizing the unconstrained objective:
\[
L_{VMPO}(\pi_\theta) = -\E_{x \sim \mu_k} \left[ \sum_a \log \pi_\theta(a |x) \psi(x, a)\right].
\]
Note that here we view $L_{VMPO}$ as another objective for approximately implementing the mirror descent policy update, which differs from the EM-algorithm perspective of \citet{song2019v}.


\subsection{Optimization landscape comparison}

Previous work of \citet{surrogate} shows that the optimization landscape corresponding to the expected-advantage objective optimized by CPO $ L_{\pi_k}(\pi_\theta)$ can exhibit suboptimal plateaus (with or without the KL constraint), and exponentially many local optima in the worst case. 
In particular, for softmax policies, Theorem 1 of \citet{surrogate} shows that even for a single observation $x\in\mathbb R^d$ 
and linear softmax activations $q_\theta(x, \cdot) = \theta x$ where $\theta \in \R^{|\cA| \times d}$, $ L_{\pi_k}(\pi_\theta)$ is non-convex in $\theta$ in general, and can have a number of local minima that is exponential in $|\mathcal{A}|$ and $d$ in the worst case.
This difficulty arises if the softmax policy is provided with an under-complete parameterization.
By contrast, if a full rank linear parameterization is used,
it is known that simple gradient descent 
with a softmax policy
will converge to a global optimum despite the non-convexity of the optimization landscape \citep{agarwaletal20,JYSzWA20}.
However, even in this over-parameterized case,
convergence can be arbitrarily slow depending on initialization \cite{meietal20}, due to the plateaus caused by softmax saturation.

These difficulties with the optimization landscape can be mitigated by the introduction of entropy regularization \cite{ahmed2019understanding,meietal20}.
Note that the objective $L_{MDPO}$ consists of $ L_{\pi_k}(\pi_\theta)$ and a KL-divergence regularizer. While this objective is also non-convex in $q_\theta$ in general, entropy regularization (included through KL divergence) can alleviate some of the issues, as empirical evidence suggests that it makes the optimization landscape more connected \cite{ahmed2019understanding}, while also mitigating the impact of softmax saturation on convergence speed \cite{meietal20}.  

The main advantage of $L_{surr}$ and $L_{VMPO}$ objectives is that they are convex in the policy optimization activations $q_\theta$. Following \citet{surrogate}, $L_{surr}$ is also calibrated for $L_{MDPO}$, and so if we successfully minimize $L_{surr}$ we also minimize $L_{MDPO}$.


\subsection{MNIST experiment}

We demonstrate the difference between the different optimization  objectives on a simple MNIST contextual bandit environment, similar to the experimental setup in \citet{surrogate} (however, we work in the online learning setting rather than batch). At each step, the environment draws a random digit image from the MNIST dataset, the agent guesses the digit, and receives a reward of 1 if the guess is correct and 0 otherwise. We train all agents for $K=100$ phases of length $\tau=1000$. We approximate value functions using feed-forward networks with a single ReLU hidden layer of size 50, and policies using feed-forward networks with two ReLU hidden layers of size 1000 and a softmax activation on the final layer. We optimize all networks using Adam with learning rate 0.0001. For each algorithm, we select the best parameter $\eta$  among  $\{10, 20, 50, 100, 200, 400, 1000\}$. We note that Surrogate and VMPO are not very sensitive to $\eta$, while MDPO and CPO perform best for $\eta \in [50, 200]$. 

The results are shown in Figure~\ref{fig:mnist}. We observe that using surrogate policy optimization objectives that are convex in softmax activations results in much faster convergence than running policy gradient on the original loss. Note that when Q-function estimates are perfect, this problem corresponds to standard supervised learning. In this case, CPO minimizes the 0-1 classification error using policy gradient, while MDPO minimizes the 0-1 error combined with a relative entropy regularizer. 
The VMPO loss corresponds to cross-entropy, the standard loss used in supervised learning, and Surrogate loss to a combination of cross-entropy and squared-error. The faster convergence of VMPO and Surrogate losses can partially be explained by the easier optimization landscape that is convex in the policy activation.


\section{Reinforcement learning experiments}

\begin{figure*}[!ht]
 \centering
\includegraphics[width=0.9\linewidth, trim=0.1cm 0cm 2.2cm 0cm, clip]{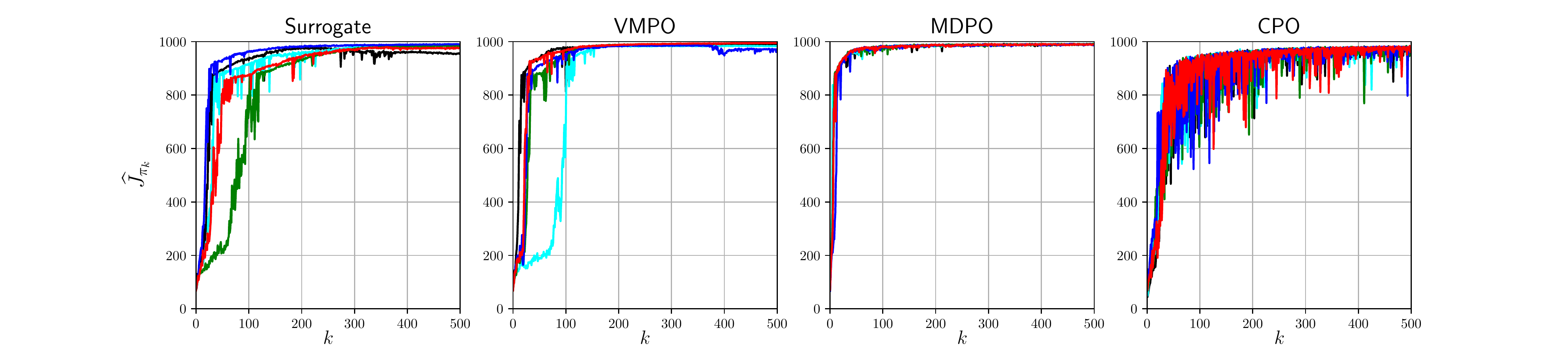}
\includegraphics[width=0.9\linewidth, trim=0.1cm 0cm 2.2cm 0cm, clip]{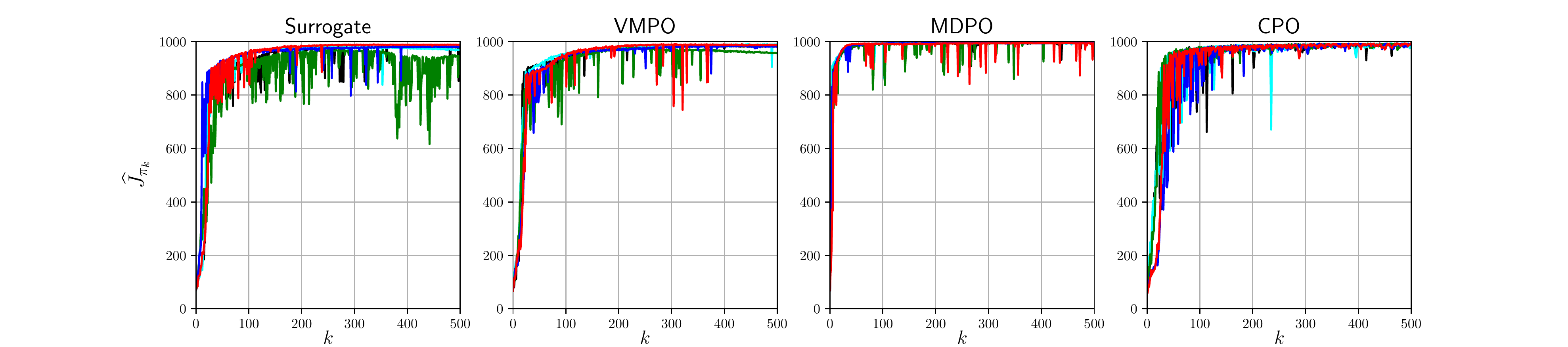}
\caption{Evaluation on the cartpole:balance environment for phase length $\tau=5000$, showing five random runs for each algorithm. Top: log-linear policies, bottom: neural network policies.}
\label{fig:cartpole5k}
\end{figure*}

 \begin{figure*}[!ht]
 \centering
\includegraphics[width=0.9\linewidth, trim=0.1cm 0cm 2.2cm 0cm, clip]{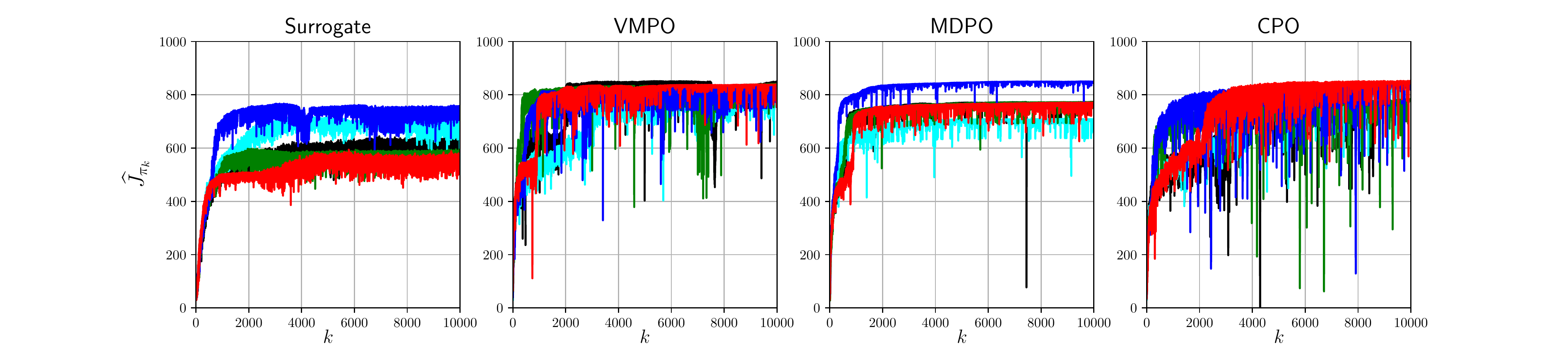}
\caption{Evaluation on the cartpole:swingup environment for phase length $\tau=10000$, showing five random runs for each algorithm. }
\label{fig:cartpole_swingup}
\end{figure*}
 
  \begin{figure*}[!ht]
 \centering
\includegraphics[width=0.9\linewidth, trim=0.1cm 0cm 2.2cm 0cm, clip]{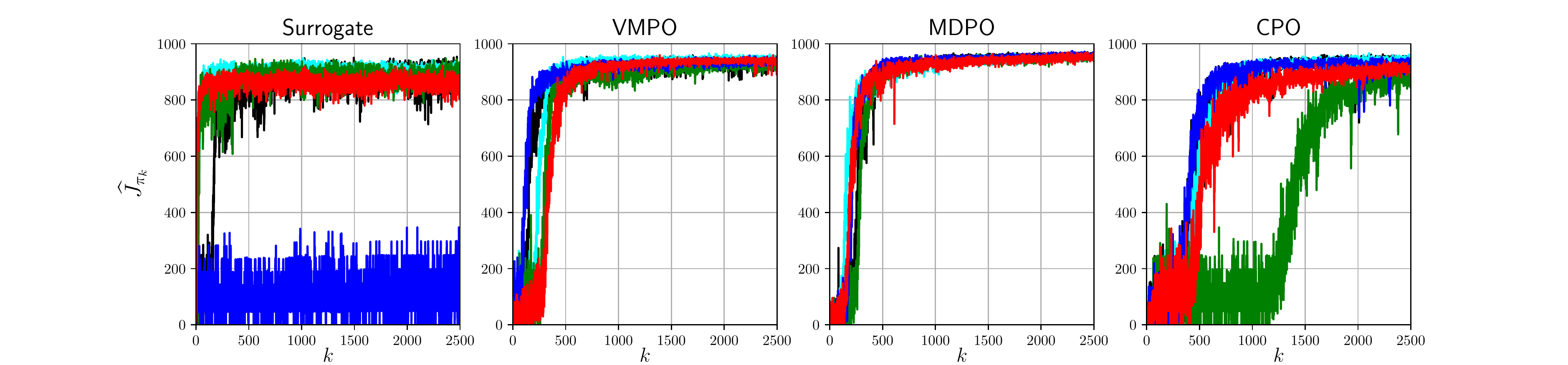}
\caption{Evaluation on the ball\_in\_cup:catch environment for phase length $\tau=20000$, showing five random runs for each algorithm.}
\label{fig:ball_in_cup}
\end{figure*}

In this section, we empirically evaluate the regularized and constrained algorithms of the previous section on reinforcement learning problems.

\paragraph{Setup.} We evaluate the algorithms on environments from the DeepMind Control Suite \cite{tassa2018deepmind}. All environments are continuous control tasks with episodes of length 1000 and rewards in $[0, 1]$. We experiment with the following environments:
\begin{itemize}
    \item \emph{cartpole:balance}: The goal is to balance an unactuated pole by applying forces to a cart at its base. The pole is nearly upright at the start of the episode. The reward function is smooth and based on the position of the pole and cart.
    \item \emph{cartpole:swingup}: The goal is to swing up and balance unactuated pole by applying forces to a cart at its base. The pole starts out pointing down. Reward is the same as in \emph{cartpole:balance}.
    \item \emph{ball\_in\_cup:catch} An actuated planar receptacle can translate in the vertical plane in order to swing and catch a ball attached to its bottom. Reward is 1 if the ball is in the cup and 0 otherwise.
\end{itemize}
We modify the environments by discretizing each action dimension to $\{-1, 0, 1\}$, and extracting multivariate Fourier-basis features from observations \cite{konidaris2011value} of order 3 (\emph{cartpole}) and 2 (\emph{ball\_in\_cup}). For \emph{cartpole:swingup}, we also include the previous 5 actions in the observation using binary indicator vectors, as this leads to faster training.

For simplicity, we use the same network architectures for all experiments. We approximate value functions using feed-forward neural networks with 50 hidden units and ReLU activations, and train them by minimizing the squared error on Monte-Carlo returns. For policies, we use feed-forward networks with layers of size $(32, 32, 32)$, ReLU activations, and a final softmax layer. For \emph{cartpole:balance}, we also experiment with log-linear policies. When implementing CPO and MDPO, similarly to previous work, we only optimize the loss on importance-weighted actions in the data; i.e. we approximate the policy expectations as 
\[
\widehat A_{\pi_k}(x_t, \pi) = \frac{\pi(a_t|x_t)}{\pi_k(a_t|x_t)}\widehat A_{\pi_k}(x_t, a_t).
\]
For VMPO and Surrogate, we use all actions. In each phase, we initialize the parameters of value functions and policies to those of the previous phase, and update them by running Adam for up to 2000 (500) steps for value functions (policies).   
For each algorithm-environment pair, we experiment with several values of the regularization / constraint parameter with $\eta \in \{0.1, 1, 5, 10, 15, 20, 25\}$, and show results for the best value. 

\paragraph{Results.} 
The results are shown in Figures~\ref{fig:cartpole5k}, \ref{fig:cartpole_swingup}, and \ref{fig:ball_in_cup}, where we plot the average reward in each phase for five random runs. 
\emph{cartpole:balance} is the easiest environment of the three; it does not require exploration, and all algorithms achieve near-optimal performance  even with log-linear policies. Figure~\ref{fig:cartpole5k} (top) shows that with log-linear policies, CPO oscillates more than the regularized algorithms, which can be explained by the optimization issues detailed in the previous sections. Interestingly, when we replace the log-linear policy by a neural net, CPO performance tends improve. One possible explanation  is that the optimization landscape issues described in the previous section are related to capacity. Indeed, the example of \citet{surrogate} showing exponentially many local optima for the expected-advantage objective involves an under-parameterized problem.  

The \emph{cartpole:swingup} and \emph{ball\_in\_cup:catch} environments are more difficult to solve, and require exploration in order to find optimal policies. Our work does not address explicit exploration, and we instead simply use longer phases in these environments in order to (randomly) discover high-reward states. The optimization issues discussed previously are more difficult to isolate in these environments due to the exploration issues. However, we do observe that although CPO finds good policies, it tends to converge more slowly than the regularized algorithms, and oscillates more when close to the optimal solution. 
Finally, we notice that optimizing the Surrogate objective can result in convergence to suboptimal policies in both environments. The only difference between Surrogate and VMPO is the addition of the squared-error loss, which often dominates the cross-entropy component. We conjecture that the squared error may lead to policies that are less greedy and take suboptimal actions more often, especially if the Q-function values for multiple actions are close in value.

\section{Discussion}

The use of KL-constrained policy updates in approximate policy iteration has become popular in recent years since the work of \citet{schulman2015trust}, and has entered the set of standard tools of RL practitioners. TRPO does lead to better performance than standard approximate policy iteration, and has been explained as a practical implementation of either CPI \citep{kakade2002approximately}, or more recently mirror descent policy optimization \citep{shani2020adaptive}. However, as we point out in this work, there is a considerable gap between theory (which suggests regularization by KL divergence) and the constrained-optimization implementation choice. 
In particular, with noisy advantages, we show that an exact implementation of TRPO is not guaranteed to converge even on simple bandit problem instances and has linear expected regret.  In addition to convergence, we show that the policy optimization landscape may be more favorable in the case of regularized updates. We hope that these observations will lead to improved practices in future applied research.

\bibliography{references}
\bibliographystyle{icml2021}

\clearpage
\onecolumn
\appendix
\onecolumn
\title{Supplementary material}

\section{Proof of Lemma~\ref{lemma:convergence}}

Let $\theta_k = \pi_k(1)$. For notation simplicity, we write $\pi_k = \pi_{\theta_k}$.
\begin{proof}
Through Pinsker's inequality,
\begin{equation*}
  2|\pi_{k+1}(1)-\theta| = \text{TV}\big(\pi_{k},\pi_{\theta}\big)\leq \sqrt{\frac{1}{2}D_{\KL}(\pi_{k}||\pi_{\theta})}\leq \sqrt{\eta/2},
\end{equation*}
where TV is the total variation distance.
Denote $\delta = \sqrt{\eta/2}$. 
Denote $T_1(\tau)$ the number of pulls for arm 1 during $\tau$ policy evaluation steps. For simplicity, we assume the policy $\pi_k$ pulls exactly $\tau\theta_k$ times for arm 1 and $\tau(1-\theta_k)$ for arm 0. At phase $k$, we compute
\begin{equation}\label{eqn:E1}
\begin{split}
    &\mathbb P\Big(\hat{r}(1)\leq\hat{r}(0)\big| \pi_k(1)=\theta_k\Big)=\mathbb P\Big(\hat{r}(0)-(-\frac{\Delta}{2})-\hat{r}(1)+\frac{\Delta}{2}\geq \Delta  \big|  \pi_k(1)=\theta_k\Big)\\
     & =  \mathbb P\Big(N(0,1)\geq \frac{\Delta}{\sigma}\sqrt{\tau\theta_k(1-\theta_k)}\Big).
\end{split}
\end{equation}

According to the update rule in Eq.~\eqref{eq:trpo2}, if $\hat{r}(1)> \hat{r}(0)$, we have
$\pi_{k+1}(1) = \min(\pi_{k}(1)+\delta, 1);$
if $\hat{r}(1)\leq \hat{r}(0)$, then $\pi_{k+1}(1) = \max(\pi_{k}(1)-\delta, 0).
$
From the assumption that $\theta_k\in[1-\delta+e, 1)$, we have
\begin{equation*}
\begin{split}
     \mathbb E[\pi_{k+1}(1)|\pi_k(1) = \theta_k] = \mathbb P\Big(\hat{r}(1)\leq\hat{r}(0)\big| \pi_k(1)=\theta_k\Big)(\theta_k-\delta)+ 1-\mathbb P\Big(\hat{r}(1)\leq\hat{r}(0)\big| \pi_k(1)=\theta_k\Big).
\end{split}
\end{equation*}
For any fixed constant $0<\delta<1$, there always exists a problem instance $\cG = (\Delta, \sigma)$ such that 
\begin{equation*}
    \mathbb P\Big(N(0, 1)\geq \frac{\Delta}{\sigma}\sqrt{\tau(1-\delta+e)(\delta+e)}\Big)\geq \frac{\delta-e}{2\delta-e},
\end{equation*}
since we can choose $\Delta$ being sufficiently small and $\sigma$ being sufficient large.
Consider the following two functions:
\begin{equation*}
    f_1(\theta) = \mathbb P\Big(N(0, 1)\geq \frac{\Delta}{\sigma}\sqrt{\tau \theta(1-\theta)}\Big), f_2(\theta) = \frac{1-\theta}{1+\delta-\theta}.
\end{equation*}
It is easy to see $f_1(\theta)$ is monotonically increasing with respect to $\theta$ as long as $\theta>1/2$ and $f_2(\theta)$ is monotonically decreasing with respect to $\theta$. Then for any $\theta_k\in[1-\delta+e, 1)$, we have 
\begin{equation*}
    \begin{split}
        \mathbb P\Big(N(0, 1)\geq \frac{\Delta}{\sigma}\sqrt{\tau \theta_k(1-\theta_k)}\Big) \geq  \mathbb P\Big(N(0, 1)\geq \frac{\Delta}{\sigma}\sqrt{\tau(1-\delta+e)(\delta+e)}\Big), \ \frac{\delta-e}{2\delta-e} \geq \frac{1-\theta_k}{1+\delta-\theta_k}.
    \end{split}
\end{equation*}
Therefore, from Eq.~\eqref{eqn:E1},
\begin{equation*}
    \begin{split}
        \mathbb P\Big(\hat{r}(1)\leq\hat{r}(0)\big| \pi_k(1)=\theta_k\Big)(1+\delta-\theta_k)\geq 1-\theta_k,
    \end{split}
\end{equation*}
which implies 
\begin{equation*}
    \mathbb P\Big(\hat{r}(1)\leq\hat{r}(0)\big| \pi_k(1)=\theta_k\Big)(\theta_k-\delta)+ 1-\mathbb P\Big(\hat{r}(1)\leq\hat{r}(0)\big| \pi_k(1)=\theta_k\Big)\leq \theta_k.
\end{equation*}

This ends the proof.
\end{proof}

\section{Proof of Lemma~\ref{lemma:trpo_regret}}

\begin{proof}
We first consider the failure probability that $\widehat \Delta_k < 0$. Let $n_k$ be the number of times arm 1 was pulled at iteration $k$. 
Recall that the reward at each pull $a_t$ is $r_t = r(a_t) + z_t$, where  $z_t \sim N(0, \sigma^2)$ i.i.d.
\begin{align*}
\mathbb P(\widehat \Delta_k < 0 |n_k = n) &= \mathbb P\left(\sum_{t=(k-1)\tau+1}^{k\tau} z_t <   ( n \Delta/2 - (\tau - n) \Delta/2)\right) \\
& =\mathbb P\left( \frac{1}{\tau} \sum_{t=(k-1)\tau+1}^{k\tau} z_t <   n \Delta / \tau - \Delta/2 \right) \\
&\geq \mathbb P\left(N(0, \sigma^2 /\tau) <  - \Delta/2 \right) \\
& = \mathbb P\left(N(0, 1) <  - \frac{\sqrt{\tau}\Delta}{2\sigma} \right)
\end{align*}
where the inequality follows since $n \Delta /\tau \geq 0$. Thus, at any iteration and for any policy $\pi_k$ and any number of pulls $n_k$, the probability of $\widehat \Delta_k < 0$ is at least $\Phi( - \frac{\sqrt{\tau}\Delta}{2\sigma})$, where $\Phi(\cdot)$ is the standard Gaussian cumulative distribution function.

Conditioned on $\Delta_k < 0$ and $\pi_k$, the next policy $\pi_{k+1}$ will be deterministic, and given by the solution of the constrained optimization problem. Let $\pi_{k+1}(1 | \Delta_k < 0, \pi_k(1))$ denote the deterministic probability of the optimal action given $\Delta_k <0$ and $\pi_k(1)$.  Note that this value will satisfy the following inequalities:
\begin{align*}
\pi_{k+1}(1 | \Delta_k < 0, \pi_k(1)) & \leq \pi_k(1) \\
\pi_{k+1}(1 | \Delta_k < 0, \pi_k(1)) & \leq \pi_{k+1}(1 | \Delta_k < 0, 1) \,.
\end{align*}
Furthermore, note that 
\begin{align*}
\pi_{k+1}(1 | \Delta_k < 0, 1) & =  \argmax_{p \in [0, 1], -\log p \leq \eta}   p \hat{\Delta}_k  
 \; = \exp(-\eta) < 1
\end{align*}
Thus, the expected regret of TRPO on this problem instance is lower-bounded by the expected number of failures and the corresponding suboptimal rewards:
\begin{align*}
\E \left[\sum_{k=1}^K \sum_{t=(k-1)\tau + 1}^{k\tau} \left( \frac{ \Delta}{2}- r_t\right) \right] 
& \geq   \frac{\tau \Delta}{2} \E \left[ \sum_{k=1}^{K-1} \mathbb I(\widehat \Delta_k < 0) (1 - \pi_{k+1}(1| \Delta_k < 0, 1)) \right] \\
& \geq  \frac{\tau \Delta}{2} \E \left[ \sum_{k=1}^{K-1} \mathbb I(\widehat \Delta_k < 0)  (1 - \exp(-\eta)) \right] \\
& \geq (K-1)  \frac{\tau \Delta}{2}\Phi\left(-\frac{\sqrt{\tau}\Delta}{2\sigma}\right)  (1 - \exp(-\eta))  \\
& =  \Phi\left(- \frac{\sqrt{\tau}\Delta}{2\sigma}\right)  \frac{ \Delta}{2} (1 - \exp(-\eta))(T-\tau).
\end{align*}

\end{proof}

\end{document}


%

%

\onecolumn
\aistatstitle{
Supplementary Materials}

\section{Policy optimization}

\paragraph{Suboptimality gap.} The loss suboptimality gap between $\pi_{k+1}(\cdot|x)$ and $u_{k+1}(\cdot|x)$ can be written as follows (dropping the dependence on $x$ for notational convenience):
\begin{align*}
   G_k &:=\eta( S_\eta^k(\pi_k) - S_\eta^k(u_k) )\\
   &= \langle \eta \widehat Q_{\pi_k}, u_{k+1} - \pi_{k+1} \rangle + D_{KL}( u_{k+1} \parallel \pi_k) - D_{KL}(\pi_{k+1} \parallel \pi_k) \\
   &= \langle \eta \widehat Q_{\pi_k} + \log \pi_k, u_{k+1} - \pi_{k+1} \rangle 
   -  \langle u_{k+1}, \log u_{k+1} \rangle 
   + \langle \pi_{k+1}, \log \pi_{k+1} \rangle\\
   &= \langle \eta \widehat Q_{\pi_k} + q_k - F(q_k) {\bf 1}, u_{k+1} - \pi_{k+1} \rangle 
  + \langle \pi_{k+1} - u_{k+1}, \log u_{k+1} \rangle 
  + \langle \pi_{k+1}, \log \pi_{k+1} - \log u_{k+1} \rangle\\
  &=\langle \eta \widehat Q_{\pi_k} + q_k - \log u_{k+1}, u_{k+1} - \pi_{k+1} \rangle 
  + \langle u_{k+1},\log \pi_{k+1} - \log u_{k+1} \rangle \\
  &= D_{KL}(u_{k+1} \parallel \pi_{k+1}) 
\end{align*}

\vfill